\documentclass[acmtog, nonacm]{acmart}

\usepackage{booktabs}
\usepackage{multirow}
\usepackage{tabularx}
\usepackage{wrapfig}
\usepackage{mathtools}
\usepackage{lipsum}
\usepackage{footmisc}
\usepackage{bbding}
\usepackage{hyperref}
\usepackage{hyperxmp}

\usepackage[ruled]{algorithm2e} 

\SetAlFnt{\small}
\SetAlCapFnt{\small}
\SetAlCapNameFnt{\small}
\SetAlCapHSkip{0pt}

\begin{document}
\def\negativevspace{}

\newcommand{\para}[1]{\vspace{.05in}\noindent\textbf{#1}}
\def\ie{\emph{i.e.}}
\def\eg{\emph{e.g.}}
\def\etal{{\em et al.}}
\def\etc{{\em etc. }}
\newcolumntype{C}[1]{>{\centering\arraybackslash}p{#1}}

\newcommand{\ourmethod}{PEGAsus }
\newcommand{\ourmethodnospace}{PEGAsus}

\newcommand{\ourop}{progressive optimization strategy }
\newcommand{\ouropnospace}{progressive optimization strategy}
\newcommand{\ouropcap}{Progressive Optimization Strategy}

\title{PEGAsus: 3D Personalization of Geometry and Appearance}

\author{Jingyu Hu}
\affiliation{%
	\institution{The Chinese University of Hong Kong}
    \country{HK SAR, China}}
\author{Bin Hu}
\affiliation{%
	\institution{The University of Hong Kong}
    \country{HK SAR, China}}
\author{Ka-Hei Hui}
\affiliation{%
	\institution{Autodesk Research}
    \country{Canada}}
\author{Haipeng Li}
\affiliation{%
	\institution{The Hong Kong University of Science and Technology }
    \country{HK SAR, China}}
\author{Zhengzhe Liu}
\affiliation{%
	\institution{Lingnan University}
    \country{HK SAR, China}}
\authornote{Corresponding author.}
\author{Daniel Cohen-Or}
\affiliation{%
	\institution{Tel-Aviv University}
    \country{Israel}}
\author{Chi-Wing Fu}
\affiliation{%
	\institution{The Chinese University of Hong Kong}
    \country{HK SAR, China}}

\begin{abstract}
We present \ourmethodnospace, a new framework capable of generating {\bf Pe}rsonalized 3D shapes by learning shape concepts at both {\bf G}eometry and {\bf A}ppearance levels.
First, we formulate 3D shape personalization as extracting reusable, category-agnostic geometric and appearance attributes from reference shapes, and composing these attributes with text to generate novel shapes.
Second, we design a \ourop to learn shape concepts at both the geometry and appearance levels, decoupling the shape concept learning process.
Third, we extend our approach to region-wise concept learning, enabling flexible concept extraction, with context-aware and context-free losses.
Extensive experimental results show that \ourmethod is able to effectively extract attributes from a wide range of reference shapes and then flexibly compose these concepts with text to synthesize new shapes.
This enables fine-grained control over shape generation and supports the creation of diverse, personalized results, even in challenging cross-category scenarios.
Both quantitative and qualitative experiments demonstrate that our approach outperforms existing state-of-the-art solutions.

\end{abstract}

\maketitle

\begin{figure}[t]
	\centering
	\includegraphics[width=0.96\columnwidth]{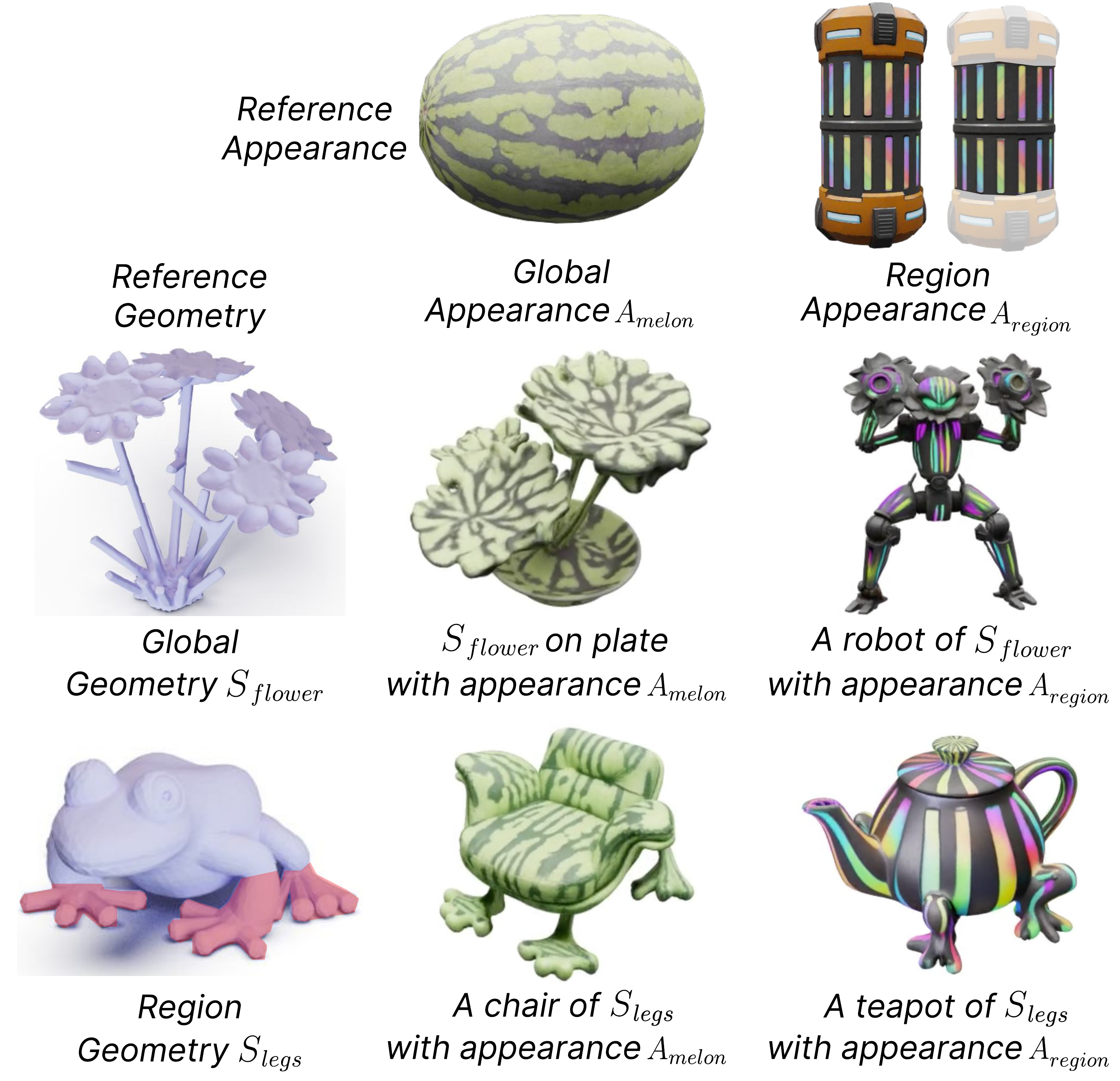}
        \vspace{-3mm}
	\caption{
        \ourmethod is capable of 
        extracting shape concepts for the global appearance of a watermelon, region-wise appearance of a stripe pattern, global geometry of a flower, and region-wise geometry of frog legs, then composing these learned concepts with text prompts in different ways to synthesize the four new shapes shown above,~\eg, a chair whose lower part follows the frog legs and appearance follows the watermelon.
     }
	\label{fig:teaser}
      \vspace{-3mm}   
\end{figure}

\vspace{-1mm}
\section{Introduction}
\label{sec:intro}
Creating distinctive 3D assets is a tedious and expensive process.
Often, significant iterations are required to create a desired 3D design. 
So, after investing huge effort in making a design, artists typically seek to reuse a design's 
geometric/appearance
attributes to create some other designs, rather than creating new ones completely from scratch.
Yet, such reusability is hard to achieve, especially for general 3D objects.
To reuse geometric/appearance attributes requires not only the capability to identify and extract geometric/appearance attributes in the reference object but also the capability to incorporate the attributes to create new designs with 
controllability.

Existing methods struggle to achieve such a goal.
Classical example-based 3D modeling methods~\cite{funkhouser2004modeling, bokeloh2010connection, chaudhuri2011probabilistic} generally reuse a reference shape by assembling retrieved parts.
This approach is essentially heuristic and operates at part level, offering limited semantic control in attribute extraction and incorporation.
3D stylization~\cite{xu2010style,li2013curve,hu_tog17} is another closely related approach, aiming to transfer the style of a reference shape to modify a given target shape.
Yet, the approach assumes a given target shape.
It {\em cannot} directly generate new shapes guided solely by the reference.
This reliance limits its usability in 3D asset design, where artists often want to create new designs, even across shapes 
of
different categories, without requiring efforts to pre-model the target.

In this work, we pursue 3D personalization.
We aim to learn attributes from references to guide the shape generation, without prescribing a target instance.
While this paradigm has been explored in image generation~\cite{gal2022image, ruiz2023dreambooth, kumari2023multi}, extending it to 3D presents unique challenges.
Unlike 2D images, where attributes are entangled, 3D asset design requires explicitly decoupling the attribute controls over geometry and appearance, while supporting region-wise concept learning and cross-category reusability; see Figure~\ref{fig:teaser} for a preview of our results.

In this paper, we introduce \ourmethodnospace, short for 3D  {\bf PE}rsonalization of {\bf G}eometry and {\bf A}ppearance (PEGA), reflecting its focus on learning reusable shape concepts 
applicable beyond a single
instance or
category.
From a reference shape, \ourmethod can effectively learn reusable and category-agnostic shape concepts, where each concept captures a specific geometric/appearance attribute.
The learned concepts can then be composed with text to synthesize new shapes, while preserving the attributes extracted from the reference.

To achieve this, we build our framework on TRELLIS~\cite{xiang2024structured}, a large 3D foundation model, which naturally decouples geometry and appearance through a two-stage shape generation pipeline. 
Leveraging this model as a robust prior, we design 
a \ourop 
to extract geometry- and appearance-level concepts from different stages of the pipeline, allowing 
their concepts to be learned independently. 
Further, \ourmethod supports both global concept learning from the entire shape and 
a
region-wise mechanism that localizes the concept learning on
a user-specified region.
For region-wise concept learning, 
we introduce two complementary objectives: 
the
context-aware loss to ensure visual coherence and 
the
context-free loss to 
isolate
the concept learning from the region's surroundings.
The learned concept is jointly represented by an optimized text embedding and a fine-tuned generator. 
At
inference, the concept embedding can be composed with 
text and fed into the fine-tuned generator to synthesize new shapes.

\ourmethod supports four personalization modes: 
\{ global or region-wise concept learning \} $\times$ \{ geometry or appearance \}. 
As Figure~\ref{fig:teaser} shows, \ourmethod 
is able to
extract the global geometry concept of the flower, the region geometry concept of the frog’s legs, global appearance concept of the watermelon, and region appearance concept of the striped pattern, and then flexibly reuse and compose these concepts with texts to synthesize different new shapes.

In summary, we make the following contributions:
\begin{itemize}
    \item We formulate the problem of \emph{3D shape personalization}
    as extracting
    reusable, category-agnostic 
    geometric and appearance attributes from reference shapes and 
    composing the attributes
    with text to generate 
    new
    shapes.

    \item We propose a concept-level 3D generation framework that independently learns and composes geometry-level and appearance-level shape concepts, while flexibly supporting both global and region-wise personalization.

    \item 
    Both quantitative and qualitative results show that 
    \ourmethod is able to produce
    a wide range of novel personalized 3D shapes with four distinct personalization modes, outperforming existing methods in both quality and flexibility.
\end{itemize}

\section{Related Work}
\label{sec:rw}

\begin{figure*}[!t]
	\centering
    \includegraphics[width=0.99\linewidth]{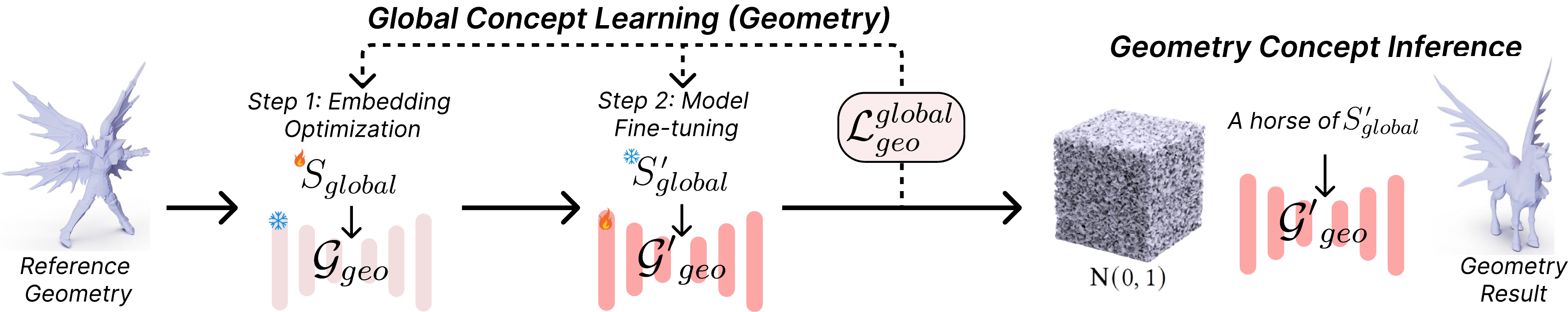}
        \vspace{-4mm}
	\caption{ 
Global concept learning in \ourmethodnospace.
Given a reference shape, 
we first optimize the text embedding $S_{global}$ to $S'_{global}$ 
then fine-tune the geometry generator $\mathcal{G}_{geo}$ to $\mathcal{G}'_{geo}$; 
both steps are guided by the global (geometry) loss $\mathcal{L}_{geo}^{global}$.
At inference, we compose the learned concept $S'_{global}$ with a text prompt and use the fine-tuned model to generate new shapes 
with the inherited
reference attributes.
This figure shows global concept learning for geometry, yet the same pipeline
can be used for appearance, 
by replacing $\mathcal{G}_{geo}$ 
with 
the
appearance generator and adjusting the reference data from geometry to appearance.
}
     \vspace{-2mm}     
	\label{fig:overview_global}
\end{figure*}

Prior related work spans several directions, 
exploring different ways of generating, editing, or adapting 3D content, 
from part-based synthesis and style propagation to large-scale generative modeling and concept learning. Below, we briefly review these areas and position our approach with respect to their assumptions and limitations.

\vspace*{1.25mm}
{\em Example-based 3D Modeling\/}~\cite{funkhouser2004modeling} aims to synthesize new shapes by copying, deforming, and assembling parts from a database of exemplar shapes.
Subsequent works~\cite{chaudhuri2011probabilistic, kalogerakis2012probabilistic, merrell2007example, fisher2012example, xu2012fit, merrell2008continuous, zhou2007terrain, bokeloh2010connection} extend this paradigm with probabilistic modeling or part-based composition.
Their reliance on manual decomposition or handcrafted priors,~\eg, symmetry, however, restricts the synthesis to a recombination of database components, 
limiting the generalization to novel or cross-category scenarios.
Recently, learning-based methods~\cite{wu2022learning, li2023patch, wu2023sin3dm, maruani2025shapeshifter} 
adopt generative models to 
implicitly extract regions or patches for producing 
plausible shape variants from a single reference shape. 
These approaches, however, focus primarily on exemplar-specific diversity,~\eg, modify the aspect ratio of the exemplar shape, offering 
limited semantic control over the generated shapes.
In contrast, our method learns reusable, category-agnostic attributes from the 
reference shape, enabling more flexible semantic control. The extracted attributes can be composed with texts to synthesize new shapes, even for cross-category scenarios.

\vspace{-1mm}
\paragraph{3D stylization}
Given a reference style input and a target content shape, 3D stylization aims to extract the stylistic patterns of the reference and re-apply them to the target to synthesize new shapes. 
This approach operates across various representations, including explicit meshes~\cite{hu_tog17,xu2010style,li2013curve,hertz2020deep,yin2021_3dstylenet,hollein2022_stylemesh, kato2018neural, michel2022text2mesh}, neural implicit fields, and Gaussian splatting~\cite{nguyenphuoc2022_snerf,huang2022_stylizednerf,liu2023_stylerf,zhang2025stylizedgs, fan2022unified, huang2021learning, kumar2023s2rf, zhang2023ref}.
To enhance the generalizability,
recent approaches~\cite{dong2024_coin3d,chen2025_artdeco, xie2024styletex} leverage strong 2D priors from large-scale image diffusion models~\cite{rombach2022high} for 3D stylization. Meanwhile, 
Qu et al.~\shortcite{qu2025_stylesculptor} employs 3D priors~\cite{xiang2024structured}
for consistent stylization across views.
Crucially, all these approaches follow a style-transfer formulation, requiring a user-provided target shape to define the geometry structure.
In contrast, our method addresses a distinct task of reference-guided 
3D
shape generation. From the reference, rather than extracting explicit geometry, we extract category-agnostic shape concepts that can be composed flexibly with 
texts
to synthesize new shapes, considering the structural semantics in the 
text, 
while preserving the learned geometric/appearance attributes from the reference shape.

\vspace*{-2mm}
\paragraph{3D Generative Modeling.}
Inspired by the success of diffusion models in 2D image and video generation~\cite{ho2020denoising,ho2022video}, recent works~\cite{hu2023_neuralwavelet, zeng2022lion, zhang20233dshape2vecset, cheng2023sdfusion, ren2024xcube, liu2023exim} 
have significantly advanced 3D generative modeling by learning compact latent representations and performing generation in the compressed space.
With the emergence of large-scale 3D datasets~\cite{deitke2023objaverse, deitke2023objaversexl}, 3D foundation models~\cite{hui2022neural,xiang2024structured,zhang2024clay,li2025triposg, zhao2025hunyuan3d, wu2025direct3d, wu2024direct3d, hong20243dtopia, he2025sparseflex, hong2023lrm} have been developed to capture strong geometric priors
supporting not only high-quality asset generation across diverse categories but also a wide range of downstream tasks,~\eg, shape analysis~\cite{du2025hierarchical} and editing~\cite{hu2024_cnsedit}.
Despite their impressive generalization, these models primarily generate samples from a learned distribution and lack mechanisms for user-specified personalization or example-based control.
In this work, we leverage the robust priors of the 3D foundation model TRELLIS~\cite{xiang2024structured} with a decoupled personalization framework to enable flexible user-specified adaptation of geometry and appearance, while maintaining the high-quality generative capabilities of the base model.

\vspace*{1.25mm}
{\em Image personalization\/} aims to adapt large text-to-image generative models~\cite{rombach2022high} to capture the visual identity of a specific subject or concept from a set of reference images, enabling faithful synthesis under new text prompts.
Textual Inversion~\cite{gal2022image} introduces personalization by learning a concept-specific text embedding, whereas DreamBooth~\cite{ruiz2023dreambooth} improves the fidelity via model fine-tuning.
Subsequent methods~\cite{kumari2023multi, ruiz2024hyperdreambooth, gu2023mix, tewel2023key, wei2023elite, shi2024instantbooth, ye2023ip, safaee2024clic} increase efficiency and controllability by updating only lightweight parameters or leveraging image encoders, and others~\cite{safaee2024clic, avrahami2023break} further extend personalization to spatially localized concepts for fine-grained, compositional control.
Despite these advances, direct 3D personalization remains underexplored.
Existing works~\cite{liu2024make, raj2023dreambooth3d, wang2024themestation,huang2025_edit360} 
propose to lift
personalized 
image diffusion priors
into 3D, so the geometry and appearance are 
entangled, with limited geometric control and weak cross-category generalization.
In contrast, we perform personalization directly in 3D, 
explicitly disentangling geometry- and appearance-level concepts, enabling cross-category concept-level shape generation.
Further, we formulate a region-wise mechanism to localize the concept learning.

\section{Overview}
\label{sec:overview}

We develop the shape concept learning pipelines in \ourmethod based on the geometry generator $\mathcal{G}_{geo}$ and appearance generator $\mathcal{G}_{app}$ of TRELLIS~\cite{xiang2024structured};
see Section~\ref{subsec: preliminary}.
Importantly, we model the shape concept {\em jointly\/} by a {\em learnable text embedding\/} 
and {\em a generator model\/} fine-tuned from $\mathcal{G}_{geo}$ or $\mathcal{G}_{app}$.
Below, we first overview the global concept learning pipeline in \ourmethod then introduce how we adapt it for region-wise concept learning.
Figure~\ref{fig:overview_global} shows the pipeline for global concept learning.
Here, we focus on geometry concept, as appearance concept follows a similar procedure.
Given a reference shape, we first adopt a \ourop (Section~\ref{ssec:global_concept}) to extract a global concept.
Then,
we optimize the text embedding $S_{global}$ to $S'_{global}$ with a frozen generator 
($\mathcal{G}_{geo}$, for the case of geometry concept) to extract
coarse
global attribute, without disrupting the generator's prior. 
Next, we fine-tune generator $\mathcal{G}_{geo}$ to $\mathcal{G}'_{geo}$ to adapt 
it
to fine-grained attribute details, while fixing the optimized text embedding $S'_{global}$. 
Crucially, both steps employ the same objective $\mathcal{L}^{global}_{geo}$, differing only in the parameters being optimized. 
At inference, the learned geometry concept embedding $S_{global}'$ can be composed with text and fed into the fine-tuned generator $\mathcal{G}'_{geo}$ to synthesize new shapes (Section~\ref{ssec:concept_usage}).

\begin{figure*}[!t]
	\centering
    \includegraphics[width=0.99\linewidth]{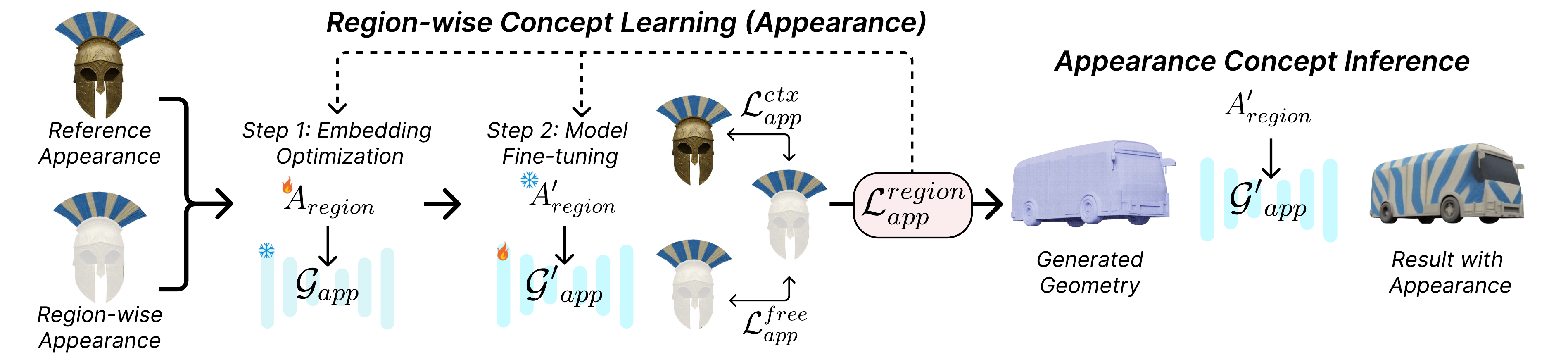}
        \vspace{-3mm}
	\caption{ 
Region-wise concept learning in \ourmethodnospace.
Given a reference appearance with a user-specified region, we perform a \ourop 
with a region-wise (appearance) objective $\mathcal{L}_{app}^{region}$.
This objective combines %
$\mathcal{L}_{app}^{ctx}$ and $\mathcal{L}_{app}^{free}$ to encourage the learned concept to be visually coherent and conceptually independent.
At inference, the 
concept synthesizes result that inherits the attributes of the specified 
region. 
While this figure shows region-wise concept learning for appearance, the pipeline applies to geometry by replacing $G_{app}$ with 
$G_{geo}$
and switching reference data to geometry.
}
     \vspace{-3mm}     
	\label{fig:overview_region}
\end{figure*}

Figure~\ref{fig:overview_region} shows the pipeline for region-wise concept learning.
For brevity, we illustrate it with appearance concept, as geometry concept follows analogously. 
Given a reference shape with a user-specified region, we first extract region-wise concept using the \ouropnospace.
Then, we adapt the global pipeline by introducing two complementary losses,~\ie, context-aware loss $\mathcal{L}^{ctx}$ and context-free loss $\mathcal{L}^{free}$, to facilitate independent yet coherent region-wise concept learning (Section~\ref{ssec:region_concept}).
At inference, the learned appearance concept can then be employed to generate shapes with the desired appearance attributes (Section~\ref{ssec:concept_usage}).

\vspace*{-1.5mm}
\section{Method}
\label{sec:method}

\subsection{Preliminary: TRELLIS}
\label{subsec: preliminary}

First, we briefly review 
TRELLIS~\cite{xiang2024structured} for the necessary background.
TRELLIS uses a unified \textit{Structured Latent} (SLAT) representation $ \{(V, Z) \mid V=\{p_{i}\}_{i=1}^{L}, Z=\{z_{i}\}_{i=1}^{L}\}$, where $V$ defines the geometry structure via spatial coordinates $p_i$ and $Z$ 
defines the associated appearance features $z_i$.
This representation decouples the 3D generation into two stages:
(i) a sparse structure stage to synthesize $V$ and (ii) a structured latent stage to generate $Z$.

\textit{Stage 1: Sparse Structure Stage.} 
This stage encodes geometry $V$ into a latent voxel grid $X$ using a pre-trained VAE, then trains the geometry generator $\mathcal{G}_{geo}$ to denoise $X$, 
conditioned on text $y$:
\begin{equation}
    \label{eq:geo_loss}
    \mathcal{L}_{geo} = \mathbb{E}_{t, X_t, y} \left[ \| v_t - \mathcal{G}_{geo}(X_t, t, y) \|^2 \right],
\end{equation}
where $X_t$ is the noisy latent at time $t$ and $v_t$ is the target velocity.
Then, the VAE decoder maps the 
denoised $X$ to generate $V$.

\textit{Stage 2: Structured Latent Stage.} 
Given the geometry $V$, the second stage trains the appearance generator $\mathcal{G}_{app}$ to denoise $Z=\{z_i\}_{i=1}^{L}$ from Gaussian noise, conditioned on text $y$ and the fixed $V$:
\begin{equation}
     \label{eq:app_loss}
    \mathcal{L}_{app} = \mathbb{E}_{t, Z_t, y, V} \left[ \| u_t - \mathcal{G}_{app}(Z_t, t, y, V) \|^2 \right],
\end{equation}
where $Z_t$ is the noisy latent at time $t$ and $u_t$ is the target velocity. The resulting structured latent $(V, Z)$ can then be decoded into various 3D representations,~\eg, 3D Gaussian splats and meshes.

Building on this SLAT representation, 
we treat geometry and appearance as distinct personalization targets, facilitating decoupled concept learning at both the geometry and appearance levels.

\subsection{Global Concept Learning}
\label{ssec:global_concept}
Fundamentally, 3D personalization involves a trade-off between (i) concept fidelity,~\ie, how accurately the concept captures the reference attributes, and (ii) transferability,~\ie, how well the concept can be composed with text to generate new shapes.
Relying solely on optimizing text embeddings is often insufficient, as the compact text embedding space lacks the capacity to encode detailed geometric or appearance attributes; see an ablation in Figure~\ref{fig:abl}(c).
Meanwhile, directly fine-tuning the generator on a single reference shape usually disrupts its 3D priors, leading to overfitting where the fine-tuned model fails to generalize to novel shapes; see Figure~\ref{fig:abl}(d). 

\paragraph{\ouropcap} 
To address the above issues, we formulate the \ouropnospace~to optimize the joint shape concept,~\ie, a text embedding and a generator model.
Below, we denote $\mathcal{G}$ as the generator, $\mathcal{D}$ the reference shape representation, and $C$ a learnable text embedding initialized from a neutral text description,~\eg, ``object'', encoded by CLIP~\cite{radford2021learning}.

\begin{itemize}
\item Step 1: Embedding Optimization.
We first capture coarse global attributes by optimizing only the text embedding.
Since the text embedding $C$ is initialized from a neutral word, we utilize a relatively high learning rate to optimize it to $C'$ while freezing the generator $\mathcal{G}$. 
This encourages the generic text embedding to align with the coarse global structure $\mathcal{D}$ without altering the generative prior.
\item Step 2: Model Fine-tuning.
To recover fine-grained details that the embedding alone cannot represent, we fix the optimized 
text embedding
$C'$ and fine-tune generator $\mathcal{G}$ to $\mathcal{G}'$ using a low learning rate. 
This conservative update alleviates overfitting to the single reference instance while allowing the model to capture high-frequency details of $\mathcal{D}$ that were missed in the embedding optimization step.
\end{itemize}

Crucially, while the two steps have different optimization parameters, the objective function remains identical.
Below, we specify the reference data $\mathcal{D}$, the learnable text embedding $C$, the generator $\mathcal{G}$, and the objective function for each case of concept learning.

\paragraph{Global Geometry Concept Learning.} For geometry concept learning, we construct the reference shape $\mathcal{D}$ by first voxelizing the reference shape, 
and encoding the voxel grid
into the latent feature grid $X$.
Then, we initialize the 
learnable text embedding
$C$ as $S_{global}$ and optimize it to $S'_{global}$.
We set the generator $\mathcal{G}$ as $\mathcal{G}_{geo}$, 
adopt the same flow matching objective as Equation~\ref{eq:geo_loss}, but replace the standard text condition $y$ with our learnable text embedding $S_{global}$:
\begin{equation}
\mathcal{L}_{geo}^{global} = \mathbb{E}_{t, X_t} \left[ || v_t - \mathcal{G}_{geo}(X_t, t, S_{global}) ||^2 \right].
\end{equation}

\paragraph{Global Appearance Concept Learning.} We define the appearance reference $\mathcal{D}$ as the SLAT representation $(V, Z)$ extracted from the reference shape, following TRELLIS~\cite{xiang2024structured},
where $Z$ serves as the target appearance features and $V$ as the fixed geometry condition. We set the generator $\mathcal{G}$ as $\mathcal{G}_{app}$, initialize the 
learnable text embedding
$C$ as $A_{global}$, and then optimize it to $A'_{global}$. The global appearance loss follows Equation~\ref{eq:app_loss}, but replacing the text prompt $y$ with the appearance text embedding $A_{global}$:
\begin{equation}
\mathcal{L}_{app}^{global} = \mathbb{E}_{t, Z_t, V} \left[ ||u_t - \mathcal{G}_{app}(Z_t, t, A_{global}, V) ||^2 \right].
\end{equation}

The final learned concept is jointly represented by the optimized embedding and the fine-tuned generator: ($S'_{global}$, $\mathcal{G}'_{geo}$) for global geometry concept and ($A'_{global}$, $\mathcal{G}'_{app}$) for global appearance concept.

\subsection{Region-wise Concept Learning}
\label{ssec:region_concept}
While global concept learning effectively captures the overall attributes of a reference shape, it lacks the precision to extract concepts from a specific local region.
To enhance user controllability and enable more flexible personalization, we introduce region-wise concept learning, which enables the extraction of geometry and appearance concepts from a user-specified region.
Given a reference shape $\mathcal{D}$ and a mask $M$ denoting a user-specified region, we learn a region-wise shape concept $(C_{region}, \mathcal{G})$, to capture the attributes within $M$.
To achieve this, we employ the same \ourop proposed in the Section~\ref{ssec:global_concept}. 

\begin{figure}[!t]
	\centering
	\includegraphics[width=0.99\linewidth]{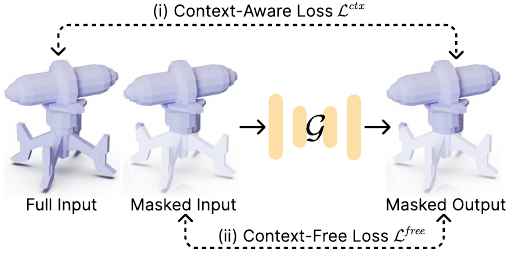}
        \vspace{-3mm}
	\caption{ 
    Region-wise concept learning objectives. (i) Context-Aware Loss takes full reference and computes loss within the masked region. (ii) Context-Free Loss uses masked input to calculate loss within the masked region.
     }
     \vspace{-4mm}
	\label{fig:illu_region}
\end{figure}

However, a challenge in region-wise concept learning is balancing context awareness with concept independence. When optimizing using the full reference shape as input while restricting supervision solely to the masked region, the generator tends to exploit the surrounding context to infer the masked region. This leads to entanglement, where the learned concept relies on the global shape rather than being an independent attribute. Conversely, if we strictly isolate the region by masking out the rest of the shape, the concept may lose its semantic coherence with the global geometry.
To address this trade-off, we introduce two complementary losses: 

\begin{itemize}
\item Context-Aware Loss ($\mathcal{L}^{ctx}$). As shown in Figure~\ref{fig:illu_region}(i), we feed the full reference shape to the generator but calculate the loss only within the mask $M$. 
The localized supervision constrains the optimization to the masked region while leveraging the global context of the reference shape.
This encourages the concept to maintain the semantic coherence between the masked region and the global structure.
\item Context-Free Loss ($\mathcal{L}^{free}$). As shown in Figure~\ref{fig:illu_region}(ii), we mask the user-specified region of the reference shape before feeding it into the generator.
By requiring the model to reconstruct the masked region in the absence of surrounding context, we encourage the learned concept to be isolated from its neighbors and thus conceptually independent. 
\end{itemize}

Below, we instantiate this region-wise concept learning framework for both geometry and appearance concepts.

\paragraph{Region-wise Geometry Concept Learning.}
For geometry, the reference data $\mathcal{D}$ refers to a latent feature grid $X$, obtained by encoding the voxelized reference shape.
Given a binary voxel mask $M$, we downsample it to align with the latent feature grid resolution, obtaining $M_{geo}$. We initialize the text embedding $C_{region}$ as $S_{region}$ and the generator $\mathcal{G}$ as $\mathcal{G}_{geo}$.
The specific losses are defined as:
\begin{equation}
\mathcal{L}_{geo}^{ctx} = \mathbb{E}_{t, X_t} \left[ || M_{geo} \odot (v_t - \mathcal{G}_{geo}(X_t, t, S_{region})) ||^2 \right]
\end{equation}
\begin{equation}
\text{and} \ \mathcal{L}_{geo}^{free} = \mathbb{E}_{t, X^{M}_t} \left[ || M_{geo} \odot (v_t - \mathcal{G}_{geo}(X^{M}_t, t, S_{region})) ||^2 \right],
\end{equation}
where $X_t$ is the noisy grid derived from the latent $X$ of the full reference shape, and $X^{M}_t$ is derived from the latent $X^M$ encoded from the masked reference shape.
\begin{equation} \mathcal{L}_{geo}^{region} = \mathcal{L}_{geo}^{ctx} + \lambda_{1} \mathcal{L}_{geo}^{free} \end{equation}

\paragraph{Region-wise Appearance Concept Learning.}
For appearance, we define the reference data as the SLAT representation $\mathcal{D} = (V, Z)$.
Given a region specified by the user, we define a binary mask $M_{app}$ on the sparse voxels $V$.
We initialize the text embedding $C_{region}$ as $A_{region}$ and the generator $\mathcal{G}$ as $\mathcal{G}_{app}$ with the following objectives:
\begin{equation}
\mathcal{L}_{app}^{ctx} = \mathbb{E}_{t, Z_t, V} \left[ || M_{app} \odot (u_t - \mathcal{G}_{app}(Z_t, t, A_{region}, V)) ||^2 \right]
\end{equation}
and
\begin{equation}
\mathcal{L}_{app}^{free} = \mathbb{E}_{t, Z^{M}_t, V^{M}} \left[ || M_{app} \odot (u_t - \mathcal{G}_{app}(Z^{M}_t, t, A_{region}, V^{M})) ||^2 \right],
\end{equation}
where $Z^M_t$ denotes the noisy latents derived from the masked appearance features $Z \odot M_{app}$, and $V^M=V \odot M_{app}$ represents the masked geometry containing only voxels within the masked region.
The total region-wise appearance objective is defined as:
\begin{equation}
\mathcal{L}_{app}^{region} = \mathcal{L}_{app}^{ctx} + \lambda_2 \mathcal{L}_{app}^{free}.
\end{equation}

The learned concept is represented by an optimized embedding and a fine-tuned generator: ($S'_{region}$, $\mathcal{G}'_{geo}$) for region-wise geometry concept and ($A'_{region}$, $\mathcal{G}'_{app}$) for region-wise appearance concept.

\subsection{Shape Concept Inference}
\label{ssec:concept_usage}
The learned concepts serve as reusable components during inference.
By leveraging the decoupled designs of geometry and appearance concepts, our framework allows them to be applied independently, enabling flexible personalization with fine-grained control.
Below, we describe the concept inference for geometry and appearance.

\paragraph{Geometry Concept Inference.} 
We employ the learned geometry concept $(S', \mathcal{G}'_{geo})$ to synthesize new 3D shapes.
Specifically, we construct a target text prompt $y_{target}$ by combining the learned text embedding $S'$ with new text prompts and feed the $y_{target}$ into the fine-tuned generator $\mathcal{G}'_{geo}$ to produce a novel shape. By leveraging the geometric attributes encoded in ($\mathcal{G}'_{geo}, S'$), we can produce a shape follows the semantic controls of the text prompt while inheriting the geometric attributes of the reference shape. 

\begin{figure}[!t]
\vspace*{-1mm}
  \centerline{\includegraphics[width=0.99\linewidth]{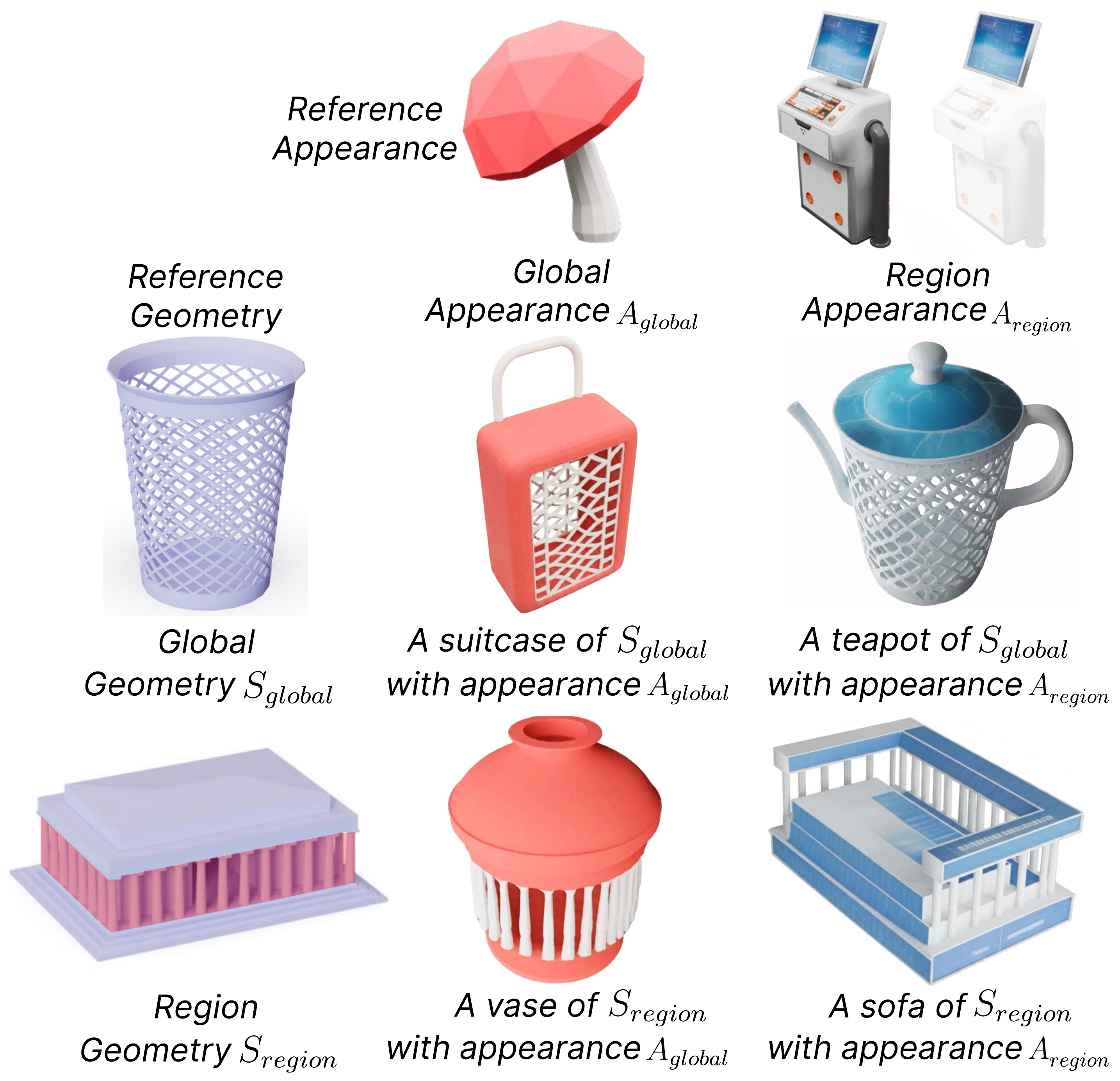}}
\vspace*{-3.5mm}
\caption{
Our method supports extracting concepts from global geometry, region-wise geometry, global appearance, and region-wise appearance, and composing the learned concepts with text to generate novel shapes.
}
\label{fig:gallery_compo}
\end{figure}

\paragraph{Appearance Concept Inference.}
Given a generated geometry $V$, we employ the learned appearance concept $(A', \mathcal{G}'_{app})$ to synthesize its appearance. The fine-tuned appearance generator $\mathcal{G}'_{app}$ produces the appearance latent $Z$, conditioned on the geometry $V$ and the appearance text embedding $A'$.
Finally, the structured latents $(V, Z)$ are decoded into a final 3D asset. By utilizing the appearance attributes encoded within $(A', \mathcal{G}'_{app})$, the produced 3D asset maintains an appearance that is consistent with the original reference.

\paragraph{Geometry/Appearance Concepts Composition.} 
As Figures~\ref{fig:teaser} and~\ref{fig:gallery_compo} show, we support composing learned geometry and appearance concepts, learned from either global or region-wise mechanism, to synthesize new shapes. We first apply the learned geometry concept composed with the text prompt to generate the shape geometry. Then, we apply the appearance concept with the generated geometry to synthesize its appearance. This design enables flexible concepts composition and provides users with fine-grained controllability.
\section{Results and Experiments}

\vspace{-1mm}
\subsection{Experiment Settings}
To evaluate the effectiveness of our method, we compare it with various baselines for both geometry and appearance concepts.

\vspace{-1mm}
\paragraph{Geometry Concept.}
To evaluate our method under this new task, we conduct a comprehensive evaluation by adapting five representative methods from closely related fields:
(i)) Qwen-Edit~\cite{wu2025qwen}, (ii) Edit360~\cite{huang2025_edit360}, (iii) StyleSculptor~\cite{qu2025_stylesculptor}, (iv) Coin3D~\cite{dong2024_coin3d}, and (v) VoxHammer~\cite{li2026_voxhammer}. 
Qwen-Edit and Edit360 perform image personalization followed by 3D reconstruction~\cite{xiang2024structured, wang2021neus}. VoxHammer targets text-guided shape editing, whereas Coin3D combines coarse geometric proxies with text for generation.
StyleSculptor requires a target shape for 3D stylization; we generate it from text using TRELLIS.
For a comprehensive evaluation, we introduce a benchmark of 30 reference shapes 
from Objaverse-XL~\cite{deitke2023objaversexl}.
Each reference is paired with three distinct text prompts targeting different categories, establishing a total of 90 evaluation cases.
More details about the geometry concept benchmark are provided in supplementary Section A.1.

\begin{table}[t]
    \begin{minipage}[t]{.49\textwidth}
        \scriptsize
        \centering
        \caption{
        Quantitative comparisons for geometry concept. Our method surpasses existing baselines across all quantitative metrics and user study results, demonstrating the effectiveness of our approach. 
        }
        \vspace{-3mm}
        \resizebox{0.99\linewidth}{!}{
        \begin{tabular}{C{2.5cm}|@{\hspace*{0.0mm}}C{0.8cm}@{\hspace*{0.0mm}}C{0.8cm}@{\hspace*{0.0mm}}C{0.6cm}@{\hspace*{0.0mm}}C{0.6cm}@{\hspace*{0.0mm}}C{0.6cm}}
        \hline 
        Method & KID $\downarrow$ & CLIP-R $\uparrow$ & QS $\uparrow$ & AS $\uparrow$ & TS $\uparrow$ \\ \hline 

        Qwen-Edit~\cite{wu2025qwen} & 0.0088 & 0.778 & 3.87 & 3.13 & 3.62 \\
        Edit360~\cite{huang2025_edit360} & 0.0082 & 0.644 & 3.43 & 2.44 & 3.53 \\
        Coin3D~\cite{dong2024_coin3d} & 0.0176 & 0.544 & 2.45 & 1.81 & 2.39 \\
        VoxHammer~\cite{li2026_voxhammer} & 0.0094 & 0.200 & 3.44 & 4.11 & 1.23 \\
        StyleSculptor~\cite{qu2025_stylesculptor} & 0.0197 & 0.589 & 3.88 & 1.30 & 4.27 \\ \hline
        Ours & \textbf{0.0064} & \textbf{0.833} & \textbf{4.41} & \textbf{4.15} & \textbf{4.31} \\
        \end{tabular}
            }
        \label{tab:geo_comp}
        \vspace{-1mm}
    \end{minipage}
    \hfill
\end{table}

\begin{table}[t]
    \begin{minipage}[t]{.49\textwidth}
        \scriptsize
        \centering
        \caption{
      Quantitative comparisons for appearance concept. Our method outperforms all baselines in both attribute preservation and overall visual quality, as demonstrated by both metric evaluations and user studies.}
      \vspace{-3mm}
        \resizebox{0.99\linewidth}{!}{
    \begin{tabular}{C{2.6cm}|@{\hspace*{0.0mm}}C{1.1cm}@{\hspace*{0.0mm}}@{\hspace*{0.0mm}}C{0.7cm}@{\hspace*{0.0mm}}C{0.7cm}@{\hspace*{0.0mm}}@{\hspace*{0.0mm}}C{0.7cm}@{\hspace*{0.0mm}}}
    \hline 
    Method & ArtFID $\downarrow$ & FID $\downarrow$ & QS $\uparrow$ & MS $\uparrow$ \\ \hline %
    
    Paint3D~\cite{zeng2024paint3d}       & 29.16 & 280.5 & 3.42 & 2.91 \\
    StyleTex~\cite{xie2024styletex}         & 30.42 & 266.5 & 2.76 & 2.25 \\
    StyleSculptor~\cite{qu2025_stylesculptor}       & 27.50 & 258.5 & 3.25 & 2.92 \\ \hline
    Ours & \textbf{23.45} & \textbf{235.9} & \textbf{4.67} & \textbf{4.37} \\
    \end{tabular}
            }
        \label{tab:app}
        \vspace{-2mm}
    \end{minipage}
    \hfill
\end{table}

\paragraph{Appearance Concept} 
To evaluate our method for the new appearance concept task,
we perform a comprehensive evaluation against three methods adapted from closely related fields:
(i) Paint3D~\cite{zeng2024paint3d}, (ii) StyleTex~\cite{xie2024styletex}, and (iii) StyleSculptor~\cite{qu2025_stylesculptor}.
Since Paint3D and StyleTex apply appearance from an image to a shape, we adapt them by rendering the reference appearance shape into an input image.
For StyleSculptor, it requires source and target images, so we render both the reference appearance shape and the geometry shape.
For a fair evaluation, we constructed a
benchmark of 30 pairs from Objaverse-XL~\cite{deitke2023objaversexl}, where each pair consists of a reference appearance shape and 
a geometry shape. More details about the appearance concept benchmark are provided in supplementary Section A.2.

\vspace{-2mm}
\subsection{Quantitative Comparison}
\paragraph{Evaluation Metrics.} 
To evaluate geometry concept, we first assess the visual quality of the generated shapes using Kernel Inception Distance (KID).
Following~\cite{raj2023dreambooth3d, liu2024make}, we utilize CLIP R-Precision (CLIP-R) to evaluate the text alignment.
For appearance concept, we adopt ArtFID~\cite{wright2022artfid} to assess appearance fidelity between the reference and generated results. Then, we follow~\cite{qu2025_stylesculptor} to use Fréchet Inception Distance (FID) to measure the visual quality of the generated appearance.
\paragraph{User Study}
Following~\cite{hu2024_cnsedit, hu2023clipxplore}, we conducted a user study with 10 participants to evaluate geometry and appearance concepts.
Participants rated results on a Likert scale from 0 (worst) to 5 (best).
For geometry, evaluations focused on: (i) Quality Score (QS), reflecting visual quality; (ii) Attribute Score (AS), assessing the preservation of geometric attributes; and (iii) Text Score (TS), evaluating how well the generated shapes follow the text.
For appearance, participants evaluated: (i) Quality Score (QS); and (ii) Matching Score (MS), measuring the similarity between the generated appearance and the reference.

Table~\ref{tab:geo_comp} presents the quantitative results for geometry concept. 
Across all metrics, our method outperforms prior approaches: it produces higher-quality geometry (higher QS, lower KID), better preserves reference attributes (higher CLIP-R, AS), and follows text control more faithfully (higher CLIP-R, TS).
Table~\ref{tab:app} reports the results for appearance concept. Our method achieves the best performance, with stronger appearance preservation (lower ArtFID, higher MS) and higher-quality generations (lower FID, higher QS).

\begin{figure}[!t]
\vspace*{-1mm}
  \centerline{\includegraphics[width=0.99\linewidth]{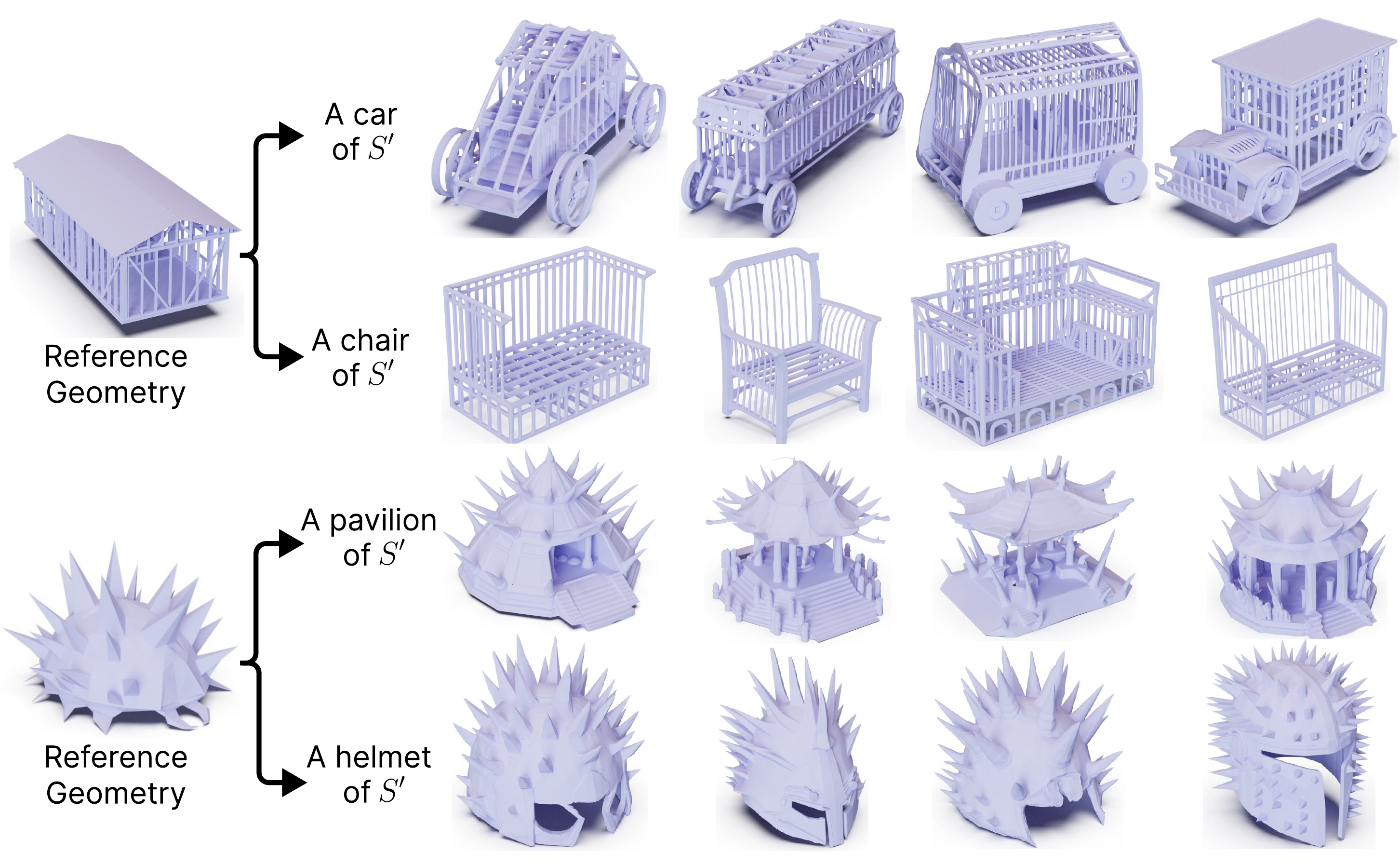}}
\vspace*{-3.5mm}
\caption{
Visual results for global geometry concept variations.
Our method generates diverse shapes that preserve the same geometric attributes,~\eg, vertical slats or spikes, while consistently following the same text controls.
}
\label{fig:gallery_geo_mul}
\end{figure}

\subsection{Qualitative Comparison}
\begin{figure*}[h]
\vspace*{-1mm}
  \centerline{\includegraphics[width=0.88\linewidth]{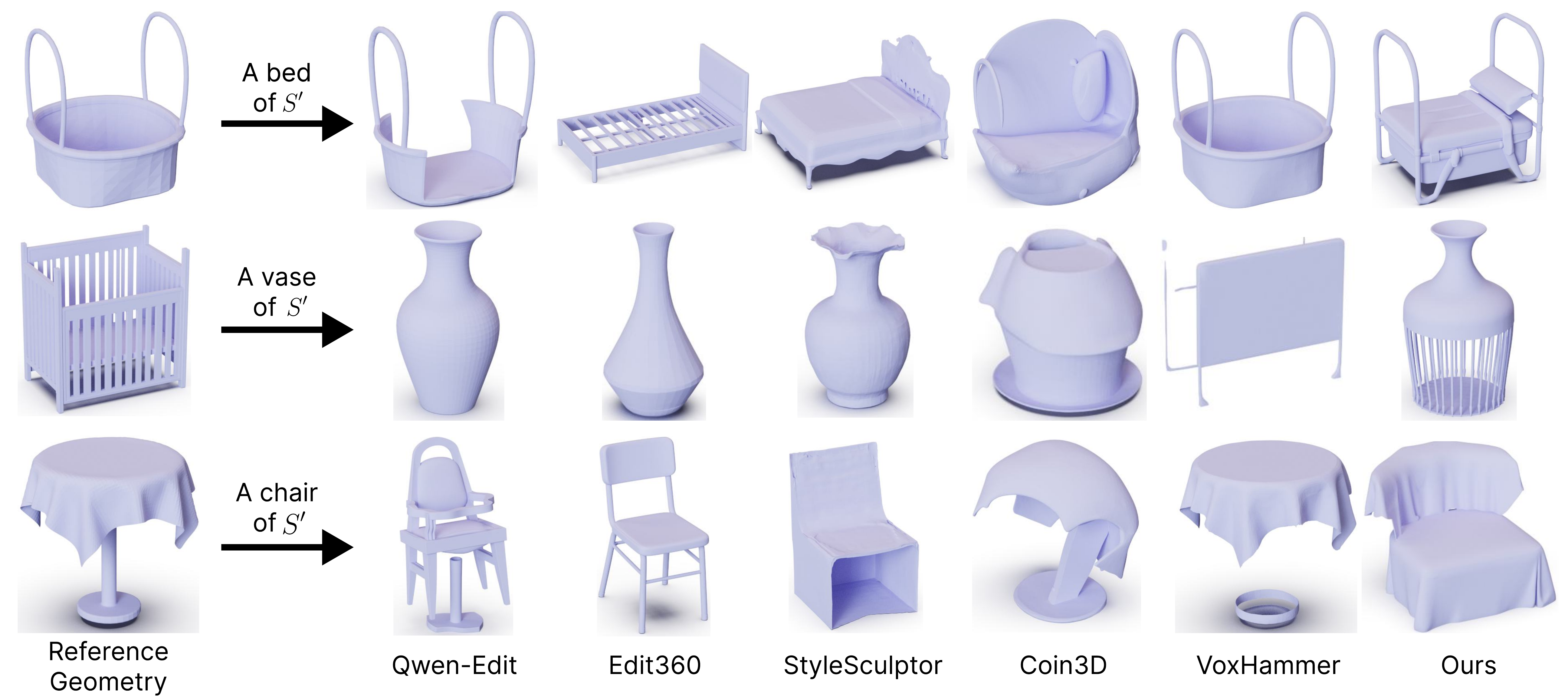}}
\vspace*{-3.5mm}
\caption{
Visual comparisons of geometry concept with Qwen-Edit~\cite{wu2025qwen}, Edit360~\cite{huang2025_edit360}, StyleSculptor~\cite{qu2025_stylesculptor}, Coin3D~\cite{dong2024_coin3d}, and VoxHammer~\cite{li2026_voxhammer}. 
Our results better preserve geometric attributes of the reference while more faithfully following text controls.
}
\label{fig:geo_comp}
\end{figure*}

\paragraph{Geometry Concept.} 
Figure~\ref{fig:geo_comp} presents visual comparisons for the geometry concept. 
As shown in second row, our method can produce results that preserve geometric attributes,~\eg, vertical slats, while adhere to the text control. In contrast, baselines either discard geometric attributes or fail to match the text.

\begin{figure*}[h]
\vspace*{-3mm}
  \centerline{\includegraphics[width=0.95\linewidth]{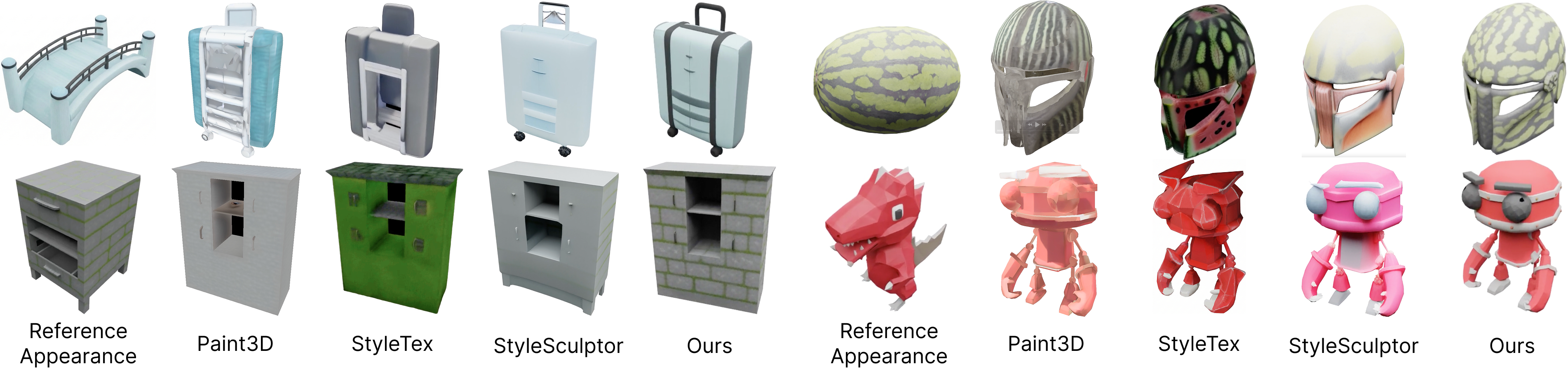}}
\vspace*{-3.5mm}
\caption{
Visual comparisons of appearance concept with Paint3D~\cite{zeng2024paint3d}, StyleTex~\cite{xie2024styletex}, StyleSculptor~\cite{qu2025_stylesculptor}. 
Our method  preserves the appearance attributes of the reference, while other approaches struggle with it.
}
\label{fig:app_comp}
\end{figure*}

\paragraph{Appearance Concept.} 
Figure~\ref{fig:app_comp} provides visual comparisons for appearance concept. Our method better preserves the appearance attribute of the reference than baselines. As shown in row 1 (left), our method retains the bridge's blue and black pattern, whereas others alter the color distribution.

More comparisons are showed in supplementary 
Sections B $\&$ C.

\subsection{More Visual Results} 
\begin{figure*}[h]
\vspace*{-1mm}
  \centerline{\includegraphics[width=0.92\linewidth]{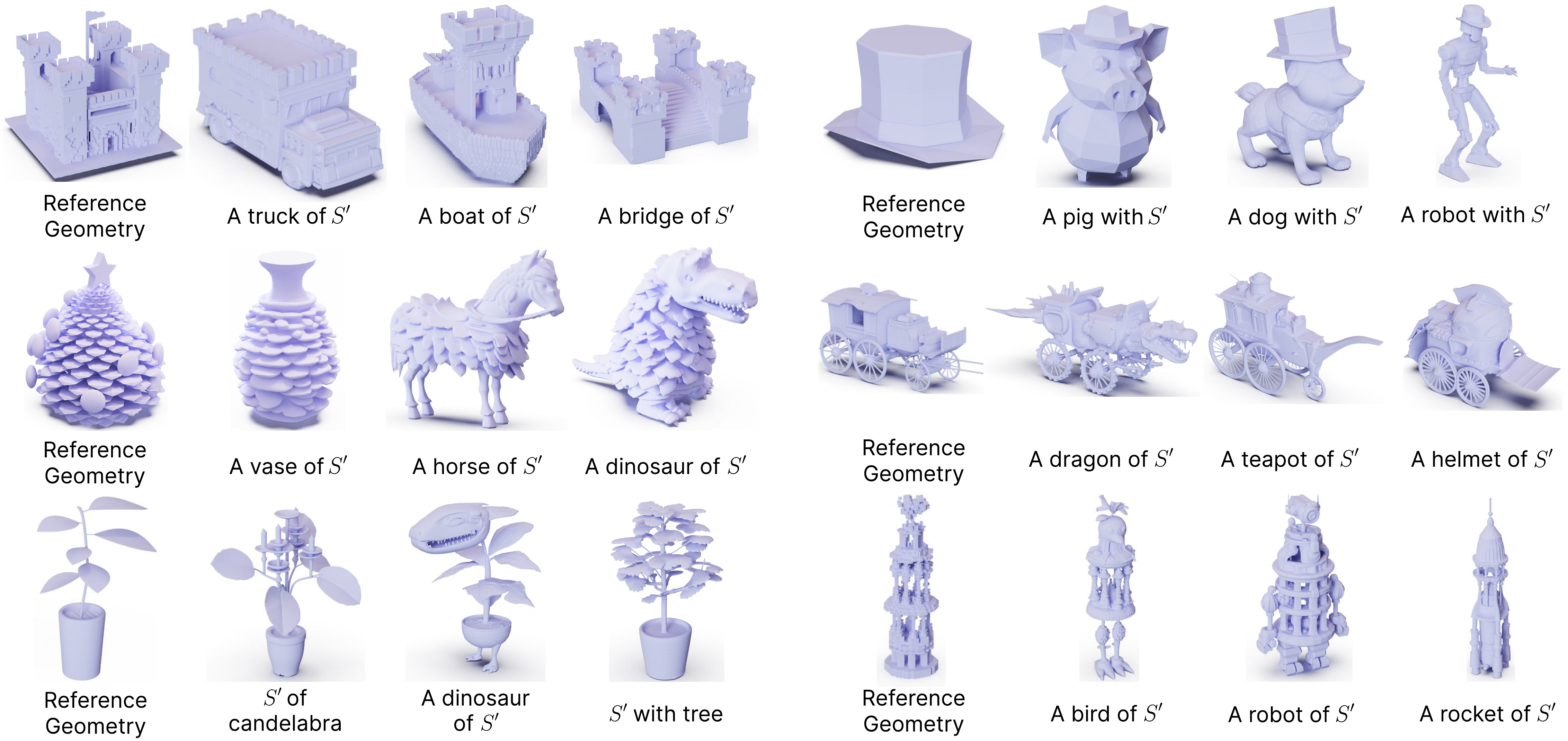}}
\vspace*{-3.5mm}
\caption{
Visual results for global geometry concept. Our method learns a global geometry concept from a reference shape and composes it with different text prompts to synthesize diverse shapes that consistently preserve the reference geometric attributes.
}
\label{fig:gallery_geo_edit}
\end{figure*}

\begin{figure*}[h]
\vspace*{-1mm}
  \centerline{\includegraphics[width=0.95\linewidth]{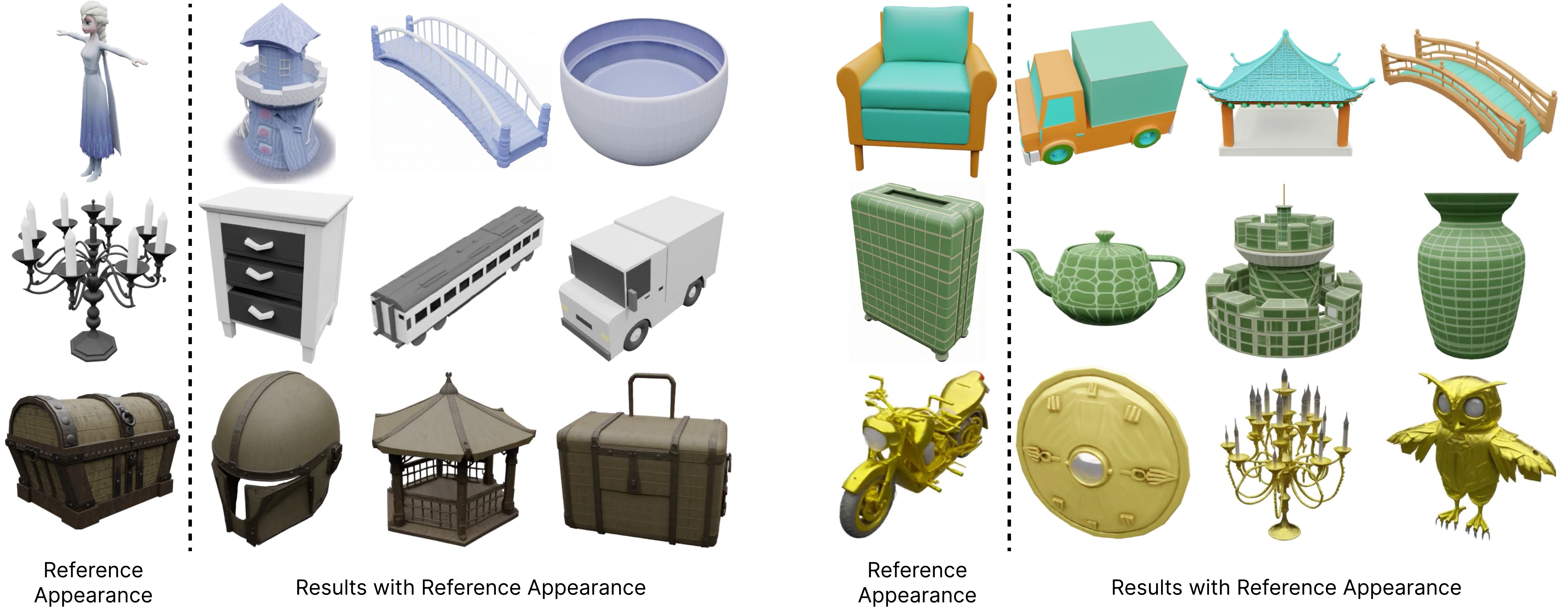}}
\vspace*{-3.5mm}
\caption{
Visual results for global appearance concept. By extracting global appearance concept, our approach produces multiple high-fidelity results that consistently reflect the reference appearance attributes.
}
\label{fig:gallery_app_trans}
\end{figure*}

\begin{figure*}[h]
\vspace*{-1mm}
  \centerline{\includegraphics[width=0.95\linewidth]{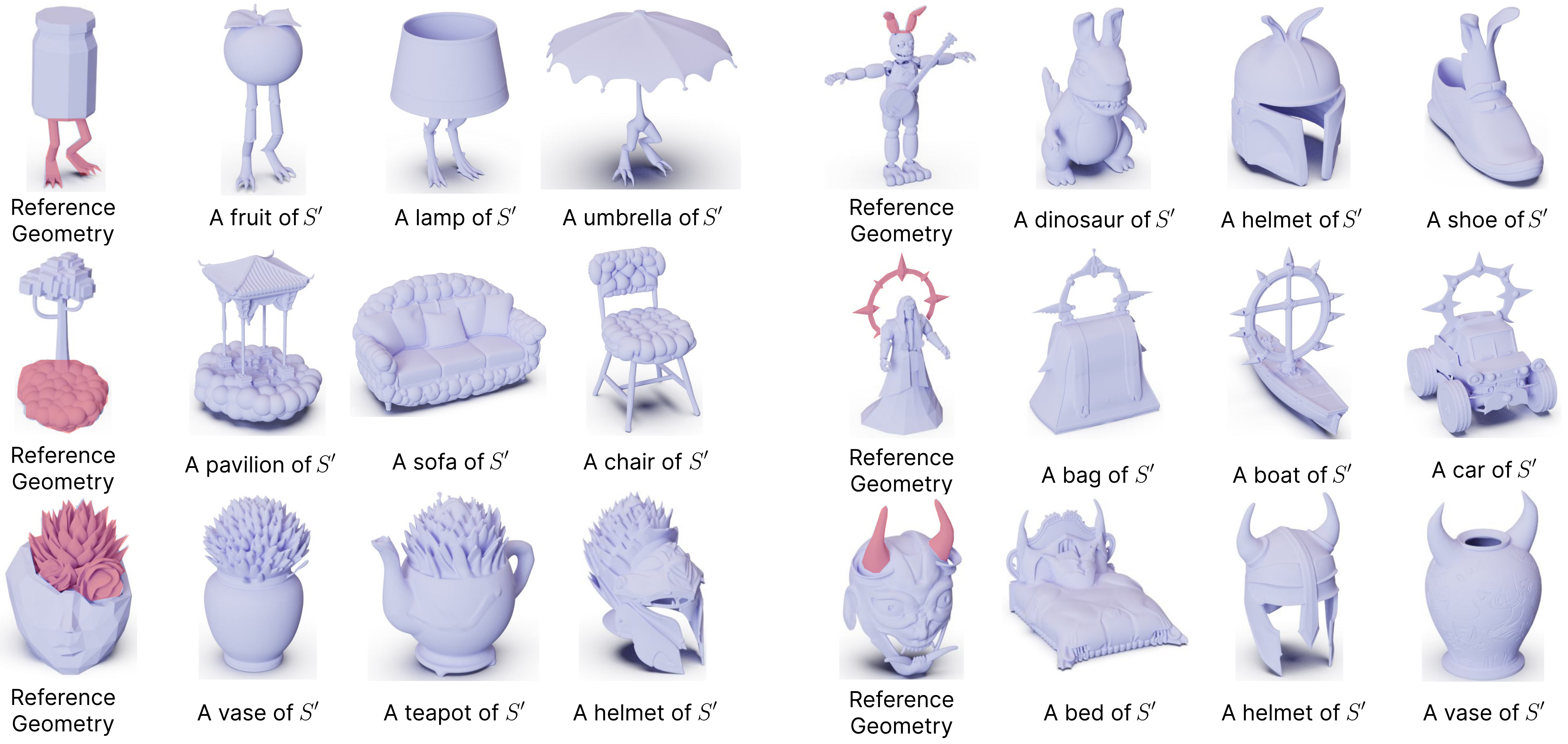}}
\vspace*{-3.5mm}
\caption{
Visual results for region-wise geometry concept. Our method enables extract the geometry concept from a user-specified region (marked in red), and compose it with text to generate novel shapes.
}
\label{fig:gallery_geo_local}
\end{figure*}

\begin{figure*}[h]
\vspace*{-1mm}
  \centerline{\includegraphics[width=0.95\linewidth]{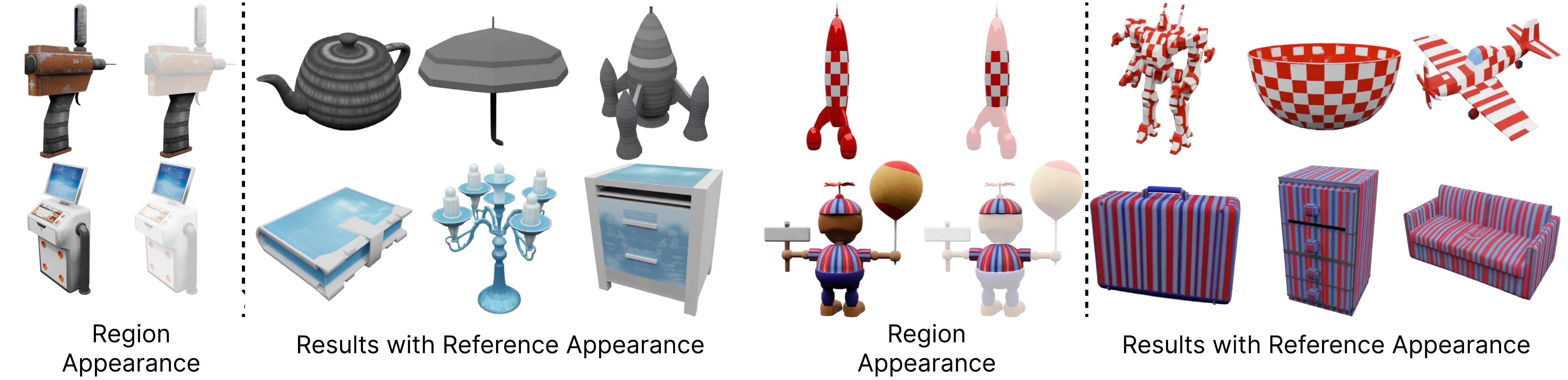}}
\vspace*{-3.5mm}
\caption{
Visual results for region-wise appearance concept. Our method enables extracting appearance concept from a localized region and reuse it to generate results that consistently inherit the reference appearance.
}
\label{fig:gallery_local_app_transfer}
\end{figure*}

We provide additional visual results for our method. We first present results for the global geometry concept: Figure~\ref{fig:gallery_geo_mul} illustrates the visual result for concept-level shape variations with the same text, while Figure~\ref{fig:gallery_geo_edit} demonstrates the synthesis of novel shapes using different text prompts.
Next, Figure~\ref{fig:gallery_app_trans} displays results for the global appearance concept. We then show results for region-wise concept: Figure~\ref{fig:gallery_geo_local} provides additional visualizations for region-wise geometry concept, and Figure~\ref{fig:gallery_local_app_transfer} presents visual results of the region-wise appearance concept. 
More visual results are provided in supplementary material Sections D, E, F, and G.

\subsection{Ablation Study}
\label{ssec:abl}
\begin{figure}[!t]
\vspace*{-1mm}
  \centerline{\includegraphics[width=0.99\linewidth]{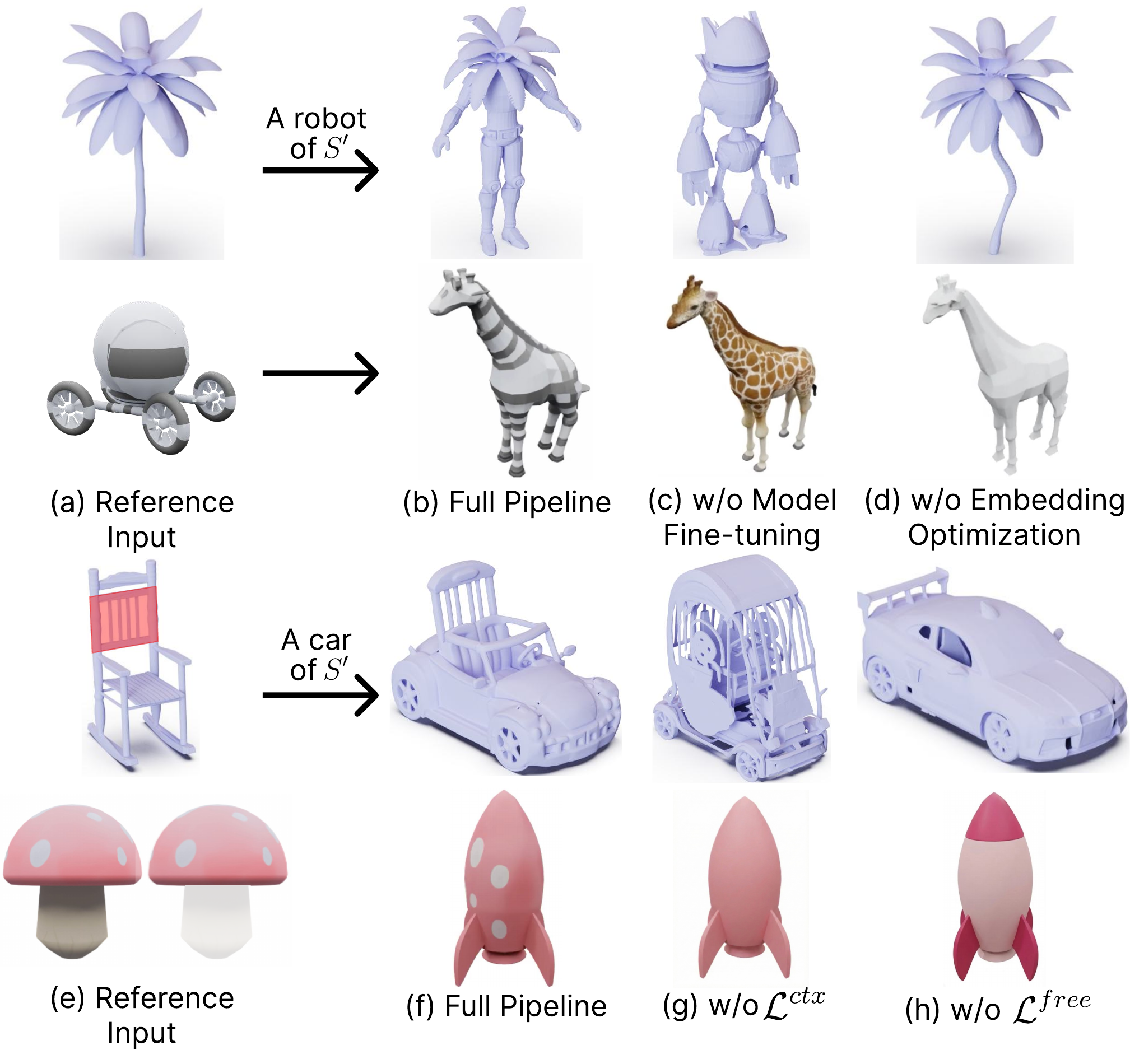}}
\vspace*{-3mm}
\caption{
Visual results for the ablation studies. (c) Removing model fine-tuning weakens attribute capture, while (d) removing embedding optimization leads to overfitting. For region-wise learning, (g) removing context-aware loss reduces visual coherence, whereas (h) removing the context-free loss hinders the model from capturing the reference attributes.
}
\vspace*{-1mm}
\label{fig:abl}
\end{figure}

We first perform ablation to validate 
\ouropnospace.
When relying solely on embedding optimization, the method struggles to capture fine-grained geometric and appearance attributes. As Figure~\ref{fig:abl}(c) shows, it fails to preserve the geometric attributes and generates a generic robot. For appearance, it cannot reproduce the black-and-white pattern, resulting in a generic giraffe-like texture.
Yet, using only model fine-tuning causes the model to overfit and lose text control. As Figure~\ref{fig:abl}(d) demonstrates, the geometry fails to follow the text control, while the appearance mimics white colors rather than the black-white patterns.
In contrast, our full pipeline effectively balances these trade-offs, ensuring both attribute preservation and text control; see Figure~\ref{fig:abl}(b).

We next ablate the two complementary losses in our region-wise objective.
First, we remove the context-aware loss $\mathcal{L}^{ctx}$ for ablation.
As shown in Figure~\ref{fig:abl} (g), removing  $\mathcal{L}^{ctx}$ degrades visual coherency. 
While the geometry concept captures vertical slats, it appears fragmented,
and the appearance concept fails to retain the pink-and-white pattern.
Second, removing $\mathcal{L}^{free}$ makes the model fail to preserve attributes.
As Figure~\ref{fig:abl}(h) shows, geometry is generated solely by the text while generated appearance struggles to maintain the pink-and-white pattern.

Table~\ref{tab:abl} shows quantitative ablation studies. Removing either embedding optimization or model fine-tuning hinders both geometry and appearance concept learning.

\begin{table}[t]
    \begin{minipage}[t]{.49\textwidth}
        \scriptsize
        \centering
        \caption{Quantitative comparison of our full pipeline against ablated cases.}
        \vspace{-3mm}
        \resizebox{0.99\linewidth}{!}{
            \begin{tabular}{C{2.5cm}|C{0.6cm}C{0.8cm}|C{0.8cm}C{0.5cm}}
            \hline

            \multirow{2}{*}{Method} & \multicolumn{2}{c|}{Geometry} & \multicolumn{2}{c}{Appearance} \\

            \cline{2-3} \cline{4-5}

             & KID $\downarrow$ & CLIP-R $\uparrow$ & ArtFID $\downarrow$ & FID $\downarrow$ \\ \hline

            Ours w/o Embedding Opt. & 0.0075 & 0.567  & 29.21 & 276.03 \\
            Ours w/o Model Fine-tuning & 0.0117 & 0.489 & 31.82 & 302.76 \\ \hline
            
            Full pipeline & \textbf{0.0064} & \textbf{0.842} & \textbf{23.45} & \textbf{235.9} \\
            \end{tabular}
        }
        \label{tab:abl}
    \end{minipage}
    \hfill
\end{table}

\section{Conclusions}

We have presented a notion of 3D personalization that shifts the focus from instance-level control to concept-level abstraction.
Rather than cloning, deforming, or editing a 
reference shape, our approach treats personalization as the extraction of reusable geometric and appearance concepts that can be recomposed across categories.
This formulation departs from target-driven pipelines that operate by modifying an existing shape, and instead enables reference-guided generation without prescribing any target instance. Conceptually, this places 3D personalization closer to language-level creativity, where abstract attributes are learned from examples and flexibly combined with semantic intent, rather than framing it as yet another geometry editing or localized shape manipulation technique. By operating 
at transferable concept level,
the method supports a more compositional and open-ended paradigm for 3D content creation.

Building on this notion, we showed that framing personalization around concept extraction 
leads to a more compositional view of 3D generation. Concepts could be learned at different scales, 
extracted
locally or globally, and combined with one another and with text in a flexible manner. Rather than producing a single edited result, this opens the door to more modular and expressive ways of creating 3D content, where new designs can emerge from the reuse and recombination of learned ideas.


\bibliographystyle{ACM-Reference-Format}
\bibliography{bibliography}


\end{document}